\documentclass[journal]{IEEEtran}
\PassOptionsToPackage{numbers}{natbib}
\usepackage[cmex10]{amsmath}
\usepackage{amssymb}
\usepackage{algorithmic}
\ifCLASSOPTIONcompsoc
 \usepackage[caption=false,font=normalsize,labelfont=sf,textfont=sf]{subfig}
\else
 \usepackage[caption=false,font=footnotesize]{subfig}
\fi
\usepackage{url}
\usepackage{graphicx}
\usepackage[table,xcdraw]{xcolor}

\usepackage{colortbl}
\usepackage{booktabs}     % table style used in books
\usepackage{siunitx}
\usepackage{multirow}     % for multi-row table
\usepackage{xcolor}
\usepackage{etoolbox}
\newrobustcmd\B{\DeclareFontSeriesDefault[rm]{bf}{b}\bfseries}
\usepackage[flushleft]{threeparttable}

\begin{document}
% \sisetup{detect-weight=true,detect-inline-weight=math} // error
\title{
Enhancing Low-Density EEG-Based Brain-Computer Interfacing with Similarity-Keeping Knowledge Distillation
}
% Improve Low-density Motor Imagery EEG Neural Networks with Knowledge Distillation

% \author{Michael~Shell,~\IEEEmembership{Member,~IEEE,}
%         John~Doe,~\IEEEmembership{Fellow,~OSA,}
%         and~Jane~Doe,~\IEEEmembership{Life~Fellow,~IEEE}}% <-this % stops a space
\author{Xin-Yao Huang, Sung-Yu Chen, and 
Chun-Shu Wei,~\IEEEmembership{Member,~IEEE}
\thanks{This work was supported in part by the National Science and Technology Council (NSTC) under Contracts 109-2222-E-009-006-MY3, 110-2221-E-A49-130-MY2, and 110-2314-B-037-061; and in part by the Higher Education Sprout Project of National Yang Ming Chiao Tung University and Ministry of Education. Corresponding author: Chun-Shu Wei (wei@nycu.edu.tw).}
\thanks{
Xin-Yao Huang, Sung-Yu Chen and Chun-Shu Wei are with the Department of Computer Science, National Yang Ming Chiao Tung University (NYCU), Hsinchu, Taiwan. Chun-Shu Wei is also with the Institute of Education and the Institute of Biomedical Engineering, NYCU, Hsinchu, Taiwan.}
}
% \thanks{J. Doe and J. Doe are with Anonymous University.}% <-this % stops a space
% \thanks{Manuscript received April 19, 2005; revised August 26, 2015.}}

% \markboth{Journal of \LaTeX\ Class Files,~Vol.~14, No.~8, August~2015}%
% {Shell \MakeLowercase{\textit{et al.}}: Bare Demo of IEEEtran.cls for IEEE Journals}


% \IEEEspecialpapernotice{(Invited Paper)}

\maketitle

\begin{abstract}
Electroencephalogram (EEG) has been one of the common neuromonitoring modalities for real-world brain-computer interfaces (BCIs) because of its non-invasiveness, low cost, and high temporal resolution.
Recently, light-weight and portable EEG wearable devices based on low-density montages have increased the convenience and usability of BCI applications. However, loss of EEG decoding performance is often inevitable due to reduced number of electrodes and coverage of scalp regions of a low-density EEG montage.
To address this issue, we introduce knowledge distillation (KD), a learning mechanism developed for transferring knowledge/information between neural network models, to enhance the performance of low-density EEG decoding.
Our framework includes a newly proposed similarity-keeping (SK) teacher-student KD scheme that encourages a low-density EEG student model to acquire the inter-sample similarity as in a pre-trained teacher model trained on high-density EEG data.
The experimental results validate that our SK-KD framework consistently improves motor-imagery EEG decoding accuracy when number of electrodes decreases for the input EEG data. For both common low-density headphone-like and headband-like montages, our method outperforms state-of-the-art KD methods across various EEG decoding model architectures. As the first KD scheme developed for enhancing EEG decoding, we foresee the proposed SK-KD framework to facilitate the practicality of low-density EEG-based BCI in real-world applications. Codes are available at https://github.com/CECNL/EEG-KD.
\end{abstract}


% \fi
\IEEEpeerreviewmaketitle

% \cite{schirrmeister2017deep,lawhern2018eegnet,wei2019spatial,li2020deep}.
\section{Introduction}
\IEEEPARstart{N}{ON}-invasive brain-computer interfacing based on electroencephalogram (EEG) has been widely developed for translating brain activities into meaningful information for a variety of applications such as communication, rehabilitation, entertainment, etc \cite{birbaumer2006breaking, mane2020bci, nijholt2008bci}.
Compared to other non-invasive neuromonitoring modalities that can facilitate a BCI, including functional magnetic resonance imaging (fMRI) \cite{sitaram2007fmri}, magnetoencephalography (MEG) \cite{mellinger2007meg}, functional near-infrared spectroscopy (fNIRS) \cite{naseer2015fnirs}, EEG outperforms in affordability, usability, and time resolution \cite{stytsenko2011evaluation, nijboer2015usability, vaid2015eeg}. 
A major weakness of EEG-based BCI is due to the limited signal-to-noise ratio, low spatial
resolution, and non-stationarity that causes suboptimal decoding accuracy \cite{vaid2015eeg,cao2011application}. 
Recently, deep-learning-based EEG decoding has attracted much attention as novel neural network models have been developed to enhance the performance of EEG decoding \cite{schirrmeister2017deep, lawhern2018eegnet, wei2019spatial, li2020deep}.

Conventionally, EEG-based BCI systems rely on laboratory-oriented high-density recording equipment to measure EEG signals \cite{liao2012biosensor}.
To set up high-density EEG recording, it requires well-trained technicians and a well-controlled environment, while bulky devices could severely limit the usability of a BCI in practical use \cite{lin2010review, ratti2017comparison}.
Recently, light-weight and portable EEG devices providing satisfactory quality of signal acquisition have been available and therefore significantly facilitate the use of EEG-based BCIs without restrictions on physical movements \cite{liao2012biosensor}.
With wearable design, dry electrodes and wireless data transmission, most of these devices are more comfortable for users, reducing the effort to set up and also expanding their usage in the real world, including education \cite{xu2018review}, stroke identification \cite{gottlibe2020stroke}, and robot control \cite{guneysu2013ssvep}.

\begin{figure}[!ht]
    \centering
    \includegraphics[width=\columnwidth]{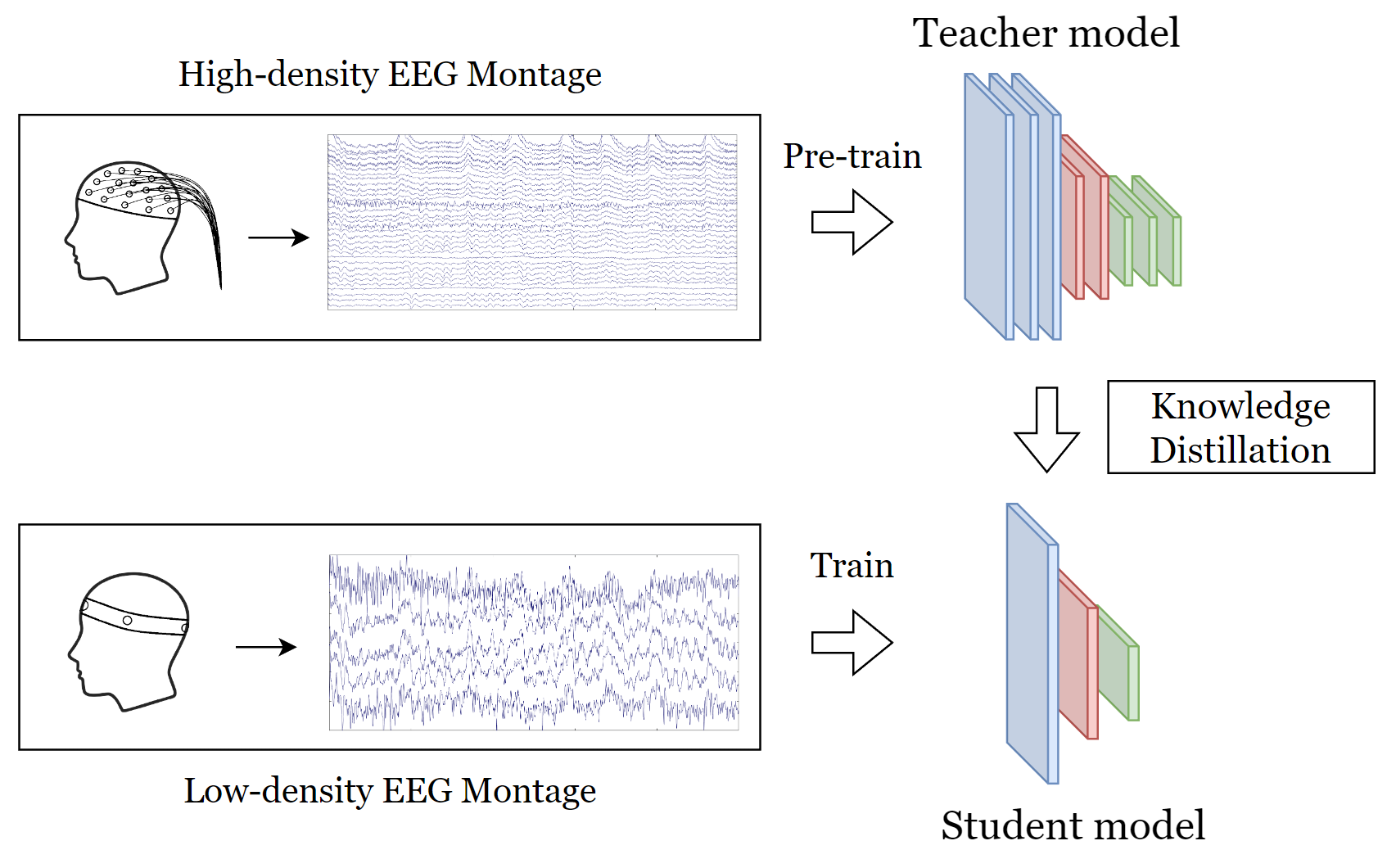}
    \caption{An illustration of the concept of using knowledge distillation (KD) to improve low-density EEG decoding.
    A pre-trained model based on EEG data recorded using high-density EEG montage serves as a 'teacher' model. A 'student' model is trained using low-density EEG data and knowledge extracted from the teacher model.
    }
    \label{fig:Intro_overview}
\end{figure}

While newly-developed low-density EEG wearable devices offer maximal convenience and comfort of use for BCI applications, there is a grand engineering challenge posed by the limited number of electrodes and recording sites.
Studies have shown that the performance of deep-learning-based EEG decoding depends on sufficient number of electrodes \cite{li2020eeg,strypsteen2021end}.
The loss in performance may be due to less information carried in signals recorded by the limited number of electrodes and from the limited coverage of scalp areas \cite{wang2012measuring,wang2016online}. 
To address the issue of low-density EEG decoding with a limited number of electrodes, we introduce knowledge distillation (KD) to improve the performance of a low-density EEG model using a pre-trained high-density EEG model, and the concept of our KD framework is illustrated as in Fig. \ref{fig:Intro_overview}. 
A pre-trained model based on EEG data recorded using high-density EEG montage serves as a 'teacher' model, and a 'student' model is trained using low-density EEG data and the knowledge extracted from the teacher model.

KD is generally applied to supporting model training using additional information extracted from an existing model. 
A deep, complex, and well-performing neural network can be 'compressed' into a shallow, small network by transferring its knowledge, such as logits output\cite{hinton2015distilling}, latent feature maps \cite{romero2014fitnets}, etc.
KD can be analogous to the interaction between a teacher and a student in the real world: A teacher can teach a student what he/she has learned.
Although various KD frameworks have been proposed and have shown their success in different fields, such as computer vision \cite{wang2021knowledge}, natural language processing \cite{sun2019patient}, and speech recognition \cite{chebotar2016distilling}, to our best knowledge, KD is rarely used in EEG-relevant studies, and none of KD methods was dedicated for EEG data modeling.
Studies have shown that improperly designed KD framework may even be harmful to the performance of a student model due to the semantic mismatch of the distilled feature maps or the capacity gap between the teacher and student networks \cite{chen2021cross, zhu2021student, ahn2019variational}.
Therefore, a KD framework dedicated for EEG data modeling is necessary to the problem of low-density EEG decoding enhancement.
We herein propose a similarity-keeping (SK) KD framework for improving low-density EEG decoding.
The proposed SK method accounts for the characteristics of deep-learning-based EEG decoding and uses relation-based KD that is capable of transferring information from a high-density EEG model to a low-density counterparts.
The main contributions of this work are three-fold: 
\begin{itemize}
\item We propose the first study that utilizes KD to improve low-density EEG decoding performance.
\item Our proposed SK framework is customized for EEG attributes and outperforms leading KD methods in low-density EEG decoding.
\item Our method significantly improves the overall performance of low-density EEG decoding regardless of the model architecture used.
\end{itemize}

%%%%%%%%%%%%%%%%%%%%%%%%%%%%%%%%%%%%%%%

\section{Related Work}

This section provides the background of KD and depicts the trend of applying KD in EEG studies.

\subsection{Background of knowledge distillation}
The original concept of KD was first introduced by Bucilua et al. \cite{bucilua2006model} for reducing the complexity of computation of ensemble learning. 
Ensemble learning is an intuitive approach to improving the accuracy of a machine learning task via fusing the predictions of multiple models trained on the same data. 
However, generating and combining predictions from individual models is often time-consuming and computationally expensive.
The method proposed in \cite{bucilua2006model} allows one single small model to be much faster but with only negligible performance loss by learning a large-scale dataset labeled by a complex model ensemble.

Later on, Hinton et al. \cite{hinton2015distilling} extended the idea to deep learning and proposed 'knowledge distillation'.
They suggested that we should identify knowledge in a trained neural network as mapping from input vectors to output vectors instead of the values of the learned parameters, and the relative probabilities of incorrect answers predicted by the cumbersome network (teacher model) can be used to train a smaller network (student model) to increase its generalizability. 
Soft scores generated by softening the distribution of the logits (input of the softmax classification output layer) can provide more information than one-hot labels in the cumbersome model, and the small network is trained to match the soft logits while predicting the true labels.
Their results showed that the performance of the student model can be successfully improved by distilling the knowledge from the teacher model. 
Subsequently, numerous studies have been conducted to find variants of KD to better define the representation of knowledge for distillation. 
In addition to Hinton's KD based on logit similarity, most KD methods can be categorized into two groups, 1) feature-based distillation using similarity of the transformation of intermediate latent feature maps, and 2) relation-based distillation using structural knowledge with the relationship between data samples \cite{gou2021knowledge}.

\subsection{Feature-based knowledge distillation}
The first feature-based KD was proposed as the FitNets \cite{romero2014fitnets} that incorporate the distance between the intermediate representation of latent feature maps, which allows the student network to be supervised by an additional regularization loss term using by the transformation of feature maps of the teacher and student networks. 
The teacher and student networks match the neural activations at the intermediate layers, providing hints to improve the training processing of the student network.
Attention transfer (AT) \cite{zagoruyko2016paying} utilizes weighted form of intermediate feature maps as the attention maps to provide information where the teacher network focuses.
Variational information distillation (VID) \cite{ahn2019variational} demonstrated that KD can be expressed as retaining high mutual information between teacher and student networks from the perspective of information theory, and the student network was trained to estimate the distribution of feature maps of the teacher network to maximize mutual information.
Activation boundary transfer (AB) \cite{heo2019knowledge} leverages activation boundaries as a useful source of knowledge using a hinge-loss-like loss without dimension reduction to keep the channel and spatial information of low-level features.

\subsection{Relation-based knowledge distillation}
Relation-based distillation leverages the relationships between the latent features generated by data samples to couple the teacher and student models.
Probabilistic knowledge transfer (PKT) \cite{passalis2018probabilistic} modeled the interactions between data samples as a probability distribution.
They employed a cosine-similarity-based function as the metric to measure the affinity between data, and KL divergence was used as the distribution divergence metric.
Correlation congruence knowledge distillation (CC) \cite{peng2019correlation} took the correlation congruence between data samples into account and minimized the pairwise correlation difference between the teacher and the student.
Relational knowledge distillation (RKD) \cite{park2019relational} revisited KD from a perspective of linguistic structuralism and focused more on the structure of data relations that can be transferred using second-order distance-wise and third-order angle-wise losses.
Similarity-preserving knowledge distillation (SP) \cite{tung2019similarity} perseved inter-sample similarities between the teacher and student models.
Contrastive representation distillation (CRD) \cite{tian2019contrastive} captured more structured representational knowledge based on contrastive learning that formulates the lower bound of mutual information between teacher and student networks.

\subsection{Knowledge distillation in EEG}
According to our survey, only a few EEG studies have explored the application of KD, and none of them have developed customized KD methods dedicated to EEG data modeling.
Sakhavi et al. \cite{sakhavi2017convolutional} applied Hinton's KD \cite{hinton2015distilling} to transfer knowledge from a teacher model trained by data of multiple subjects for reducing the calibration time for a new subject.
Zhang et al. \cite{zhang2021distilling} adopted Hinton's KD and SP \cite{tung2019similarity} to compress a cumbersome capsule network by defining knowledge for low-level and high-level capsules for EEG-based affective computing.
Wang et al. \cite{wang2021fldnet} used Hinton's KD \cite{hinton2015distilling} to facilitate multi-step model training in a multi-layer LSTM model for EEG-based emotion recognition.
Joshi et al. \cite{joshi2021deep} combined AT and Hinton's KD to conduct cross-modal knowledge distillation from an EEG-based teacher model to an ECG-based student model for sleep staging.

\begin{table}[ht!]
\centering
    \resizebox{\columnwidth}{!}{%
\begin{tabular}{c|c|c}
\toprule
Method  & Knowledge to distill & Loss function    \\ \toprule
FitNet  & Feature maps          & MSE \\ \midrule
AT      & Attention maps        & MSE \\ \midrule
VID     & Mutual information    & Variation lower bound \\ \midrule 
AB      & Activation boundaries & Alternative hinge loss \\ \bottomrule
\end{tabular}%
}
    \caption{Feature-based knowledge distillation}
    \label{tab:Feature_base}
\end{table}

\begin{table}[ht!]
\centering
    % \resizebox{\columnwidth}{!}{%
% \begin{tabular}{c|c|c|c}
% Method  & Knowledge to distill & Loss function & Similarity criterion   \\ \toprule
% SP      & Inter-sample similarity   & MSE & Dot product \\ \midrule
% CC      & Inter-sample similarity   & MSE & Kernel method \\ \midrule
% RKD     & Inter-sample relationship & Distance-wise / Angle-wise loss & - \\ \midrule
% PKT     & Feature distribution      & KL-divergence & Cosine similarity \\ \midrule
% CRD     & Mutual information        & Variational lower bound & Cosine similarity \\ \bottomrule
% \end{tabular}%
% }

\resizebox{\columnwidth}{!}{%
\begin{tabular}{c|c|c}
\toprule
Method  & Knowledge to distill & Loss function \\ \toprule
PKT     & Feature distribution      & KL-divergence \\ \midrule
CC      & Inter-sample similarity   & MSE \\ \midrule
RKD     & Inter-sample relationship & Distance-wise / Angle-wise loss \\ \midrule
SP      & Inter-sample similarity   & MSE \\ \midrule
CRD     & Mutual information        & Variational lower bound\\ \bottomrule
\end{tabular}%
}
    \caption{Relation-based knowledge distillation}
    \label{tab:Relation_base1}
\end{table}

\begin{table}[ht!]
\centering
    \resizebox{\columnwidth}{!}{%
\begin{tabular}{c|c|c}
\toprule
Research  & Application of KD      & KD method    \\ \toprule
Sakhavi et al.    & Calibration time reduction    & Hinton's KD \\ \midrule 
Zang et al.      & Model compression             & Hinton's KD + SP \\ \midrule
Wang et al.    & Ensemble model training      & Hinton's KD \\ \midrule
Joshi et al.   & Cross modal transfer          & Hinton's KD + AT \\ \bottomrule
\end{tabular}%
}
    \caption{Applications of knowledge distillation in EEG studies}
    \label{tab:Relation_base2}
\end{table}

%%%%%%%%%%%%%%%%%%%%%%%%%%%%%%%%%%%%%%%

\section{Our Approach}

\begin{figure*}
    \centering
    \includegraphics[width=\textwidth]{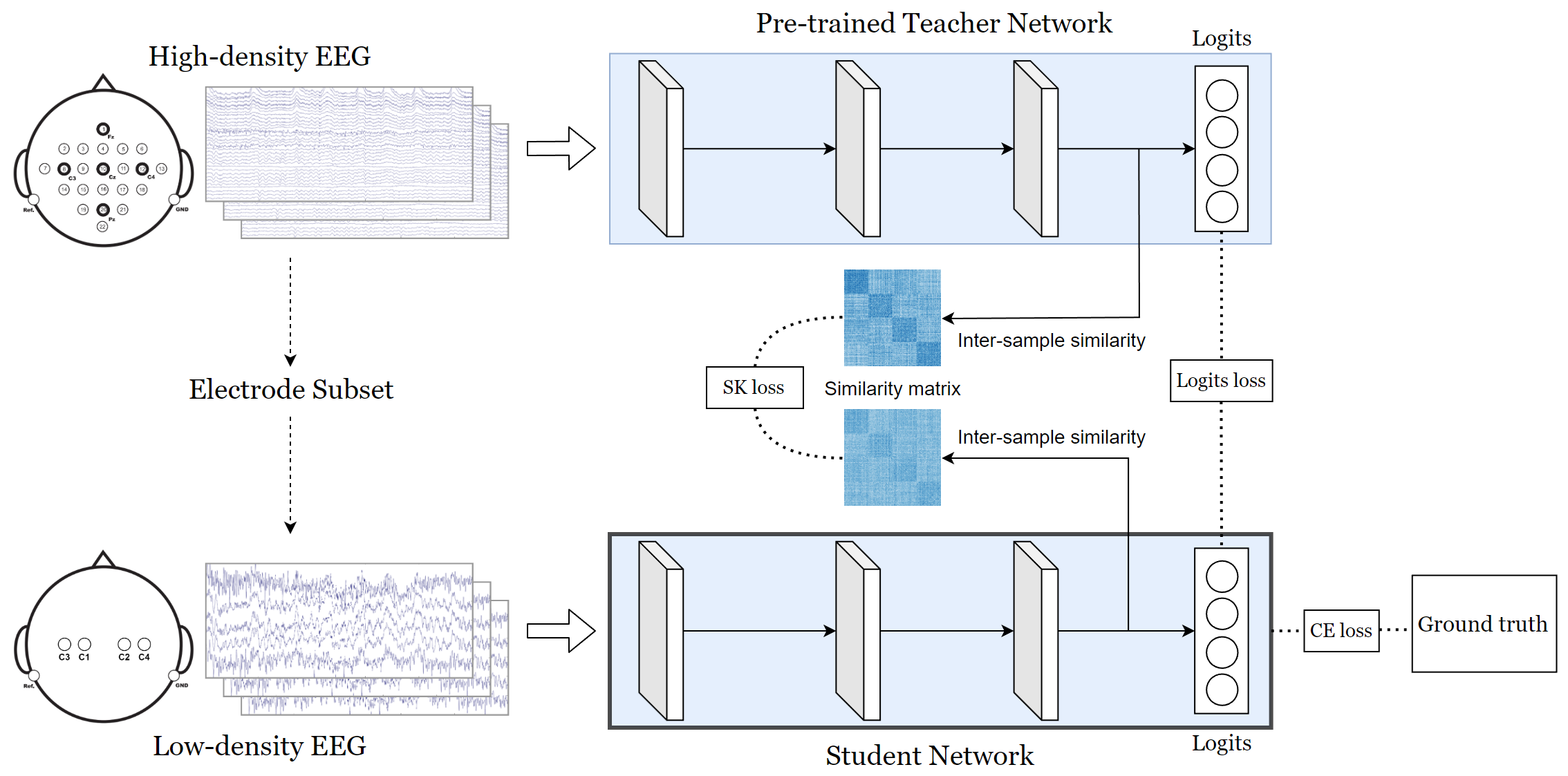}
    \caption{An overview of the proposed framework using similarity-keeping knowledge distillation (SK-KD). 
    A pre-trained 'teacher' network model using high-density EEG data supports the knowledge distillation for the 'student' network model based on 1) a similarity-keeping (SK) loss function that estimates the distance between the inter-sample similarity matrices of the two networks and 2) the logits loss function that estimates the similarity in their logtis (soft classification outputs). 
    Utilizing the cross-entropy (CE), SK, and logits loss functions, the student network model is then trained by the low-density EEG data with a subset of electrodes selected from the high-density electrodes.
    The upper montage is excerpted from \cite{brunner2008bci}.
    }
    \label{fig:Overview}
\end{figure*}

\begin{figure}
    \centering
    \includegraphics[width=\columnwidth]{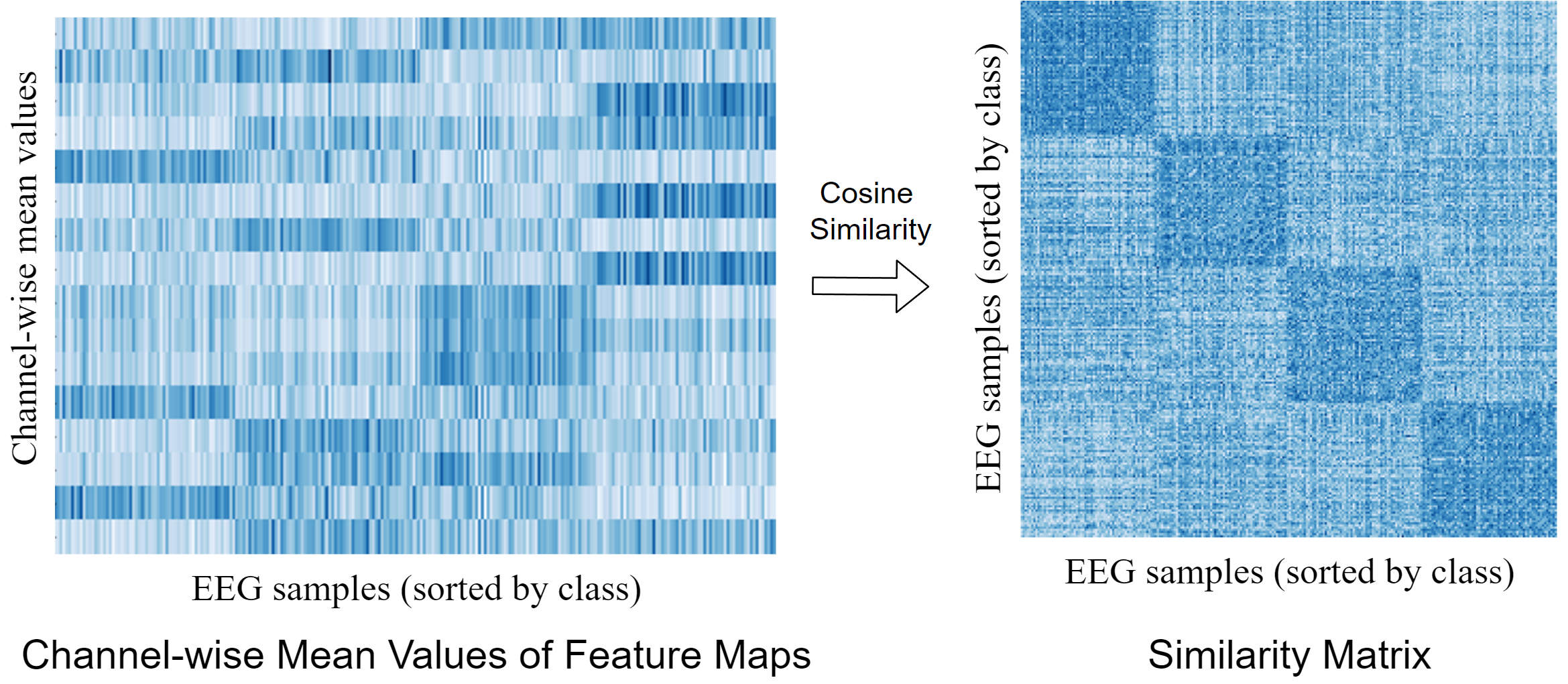}
    \caption{
    Channel-wise average feature maps for each input samples and similarity matrix of a trained EEGNet. 
    We use feature maps at the penultimate layer of the network, and entries of similarity matrix is calculated by the mean of channel-wise zero-centered cosine similarity between feature maps of two input samples, representing the similarity between them.
    Input samples are sorted by class, and similarity matrix is L2-normalized for better visualization.
    Dark blue marks higher values.
    We can see that samples in the same class tend to generate similar feature map patterns and get high inter-sample similarity scores.
    }
    \label{fig:Activation_Mat}
\end{figure}

The proposed SK framework is illustrated as in Fig. \ref{fig:Overview}. 
The components in the framework consist  of a pre-trained teacher model, a target student model, and the scheme of knowledge distillation. 
A teacher model is pre-trained in advance using high-density EEG data. To enhance the performance of the student model for a target low-density EEG montage, the student network model is trained using a specific subset of high-density EEG data based on a certain montage corresponding to a target low-density device.
During the training process, cross-entropy loss, logits loss (as in Hinton's KD), and the proposed SK loss functions are integrated and adopted to improve the performance of the student network model by distilling the additional information of classification score and inter-sample similarity using the pre-trained teacher network model.

\subsection{Similarity-kept knowledge distillation}
The main characteristic of the proposed SK framework is the proposed SK loss function that enhances the training process of the student network using the relation between EEG input samples. 
As suggested in \cite{tung2019similarity}, the knowledge of inter-sample relation can be represented as a similarity matrix, which is transferred through minimizing the distance between the similarity matrices of the teacher and student networks. 
As shown in Fig. \ref{fig:Activation_Mat}, the similarity between every two latent features within a mini-batch during training forms a similarity matrix, which is used to compute the SK loss function.
Our processing flow to acquire the SK loss is tailored by the consideration of the nature of EEG and its representation within an EEG decoding model.
Due to the non-stationarity of EEG \cite{cao2011application}, the mean level of each input EEG sample may vary, and thus we appled centering to the latent feature corresponding to each input EEG sample as a procedure of zero-mean normalization. 
Meanwhile, EEG recording is well known to be susceptible to artifact and noise \cite{winkler2015influence}. 
We chose cosine similarity to estimate the similarity between the latent features corresponding to the input samples, as the cosine similarity is a distance metric less sensitive to unstable bursts of magnitudes in the EEG feature maps.
In sum, the proposed SK loss term provides a robust estimation of inter-sample similarity for the EEG latent feature maps, and encourages the student network to match the inter-sample relation in the teacher network.

Given a mini-batch whose size is $N$, and the SK loss function for our similarity-keeping knowledge distillation is defined as:
\begin{equation}
    L_{SK}=\sum\limits_{(l,l')}mse(M^{(l)}_{T},\ M^{(l')}_{S}),
\label{eq:SK}
\end{equation}
where $mse$ denotes mean square error function, and $(l, l')$ denotes the pair of the teacher-student layer, and $M^{(l)}_T\in R^{N \times N}$ and $M^{(l')}_S \in R^{N \times N}$ denote the similarity matrices for the teacher and student networks at layer $l$ and layer $l'$, respectively.

As described above, we use the mean of channel-wise zero-centered cosine similarity as the metric to construct similarity matrices $M^{(l)}_T$ and $M^{(l')}_S$ with entries representing the similarity between input samples.
Let $F^{(l)}_T \in R^{N \times C \times H \times W}$ and $F^{(l')}_S \in R^{N \times C' \times H' \times W'}$ denote the feature maps generated by the teacher network at layer $l$ and the feature maps generated by the teacher network at layer $l'$ for a mini-batch, respectively. 
We first zero-center both feature maps along their channel dimension:
% Let $F^{(l)}_T \in R^{N \times C \times H \times W}$ denote the feature maps generated by the teacher network at layer $l$, and $F^{(l')}_S \in R^{N \times C' \times H' \times W'}$ denote the feature maps generated by the student network at layer $l'$, we first zero-center both feature maps along their channel dimension:
\begin{equation}
\begin{aligned}
    \hat{F}^{(l)}_{T[:,c]} &= F^{(l)}_{T[:,c]} - mean(F^{(l)}_{T[:,c]}) \\
    \hat{F}^{(l')}_{S[:,c]} &= F^{(l')}_{S[:,c]} - mean(F^{(l')}_{S[:,c]}),
\end{aligned}
\label{eq:mean}
\end{equation}
where $\hat{F}^{(l)}_{T} \in R^{N \times C \times H \times W}$ and $\hat{F}^{(l')}_{S} \in R^{N \times C' \times H' \times W'}$ denote the zero-centered feature maps and $[:,c]$ denotes the $c$-th channel for each sample in the feature maps.
Then the entries of $M^{(l)}_{T}$ and $M^{(l')}_{S}$ are calculated by:
\begin{equation}
\begin{split}
\begin{aligned}
    M^{(l)}_{T[i,j]} &= \frac{1}{C} \sum_{c=1}^{C} cos 
    \langle \hat{F}^{(l)}_{T[i,c]}\ ,\hat{F}^{(l)}_{T[j,c]} \rangle \\
    M^{(l')}_{S[i,j]} &= \frac{1}{C'} \sum_{c=1}^{C'} cos 
    \langle \hat{F}^{(l')}_{S[i,c]}\ ,\hat{F}^{(l')}_{S[j,c]} \rangle,
\end{aligned}
\end{split}
\label{eq:cosine}
\end{equation}
where $M^{(l)}_{T[i,j]}$ and $M^{(l')}_{S[i,j]}$ denote the entries indexed by $[i,j]$ in the similarity matrices, $C$ and $C'$ denote the number of channels in each feature map, $cos$ denotes the cosine similarity, $\hat{F}^{(l)}_{T[i,c]}$ and $\hat{F}^{(l)}_{T[i,c]}$ denote the $c$-th channel in $\hat{F}^{(l)}_{T[i]}$ and $\hat{F}^{(l)}_{T[i]}$ which are the features generated by the $i$-th sample and $j$-th sample with the teacher network, and $\hat{F}^{(l')}_{S[i,c]}$ and $\hat{F}^{(l')}_{S[i,c]}$ denote the $c$-th channel in  $\hat{F}^{(l')}_{S[i]}$ and $\hat{F}^{(l')}_{S[i]}$ which are the features generated by the $i$-th sample and $j$-th sample with the student network.
The total loss function to train the student network model with the SK loss is defined as follows:
\begin{equation}
\begin{split}
\begin{aligned}
    Loss = L_{CE}(y, \sigma(z_{S})) + \beta\cdot L_{SK}
\end{aligned}
\end{split}
\label{eq:Loss}
\end{equation}
where $L_{CE}$ denotes the cross-entropy function, y is the one-hot label of ground truth, $\sigma(\cdot)$ denotes the softmax function, $z_{S}$ is the output logit vector of the student model, and $\beta$ denotes the balancing hyperparameter.

\subsection{Hinton's knowledge distillation}
Hinton's KD \cite{hinton2015distilling} aims to compress the teacher network by encouraging the student network to generate predictions similar to those of the teacher model while predicting true labels.
In other words, the student network is trained under the supervision of the ground truth and the logits (input class of softmax classification layer) of the teacher network at the same time.
Since the probability distribution would be dominated by one class, one hyperparameter temperature $T$ is often introduced to soften logits output.
By raising the temperature, we can soften the predicted probability distribution and increase the weights of those tiny prediction scores in the loss function to transfer the information between their relationship.
The relationship between these tiny prediction scores is also called dark knowledge and can be used to facilitate the training of the student network \cite{allen2020towards}.
The total loss function for training the student network model with logits loss is given by:
\begin{equation}
\begin{split}
\begin{aligned}
    L_{KD}=&(1-\alpha) \cdot L_{CE}(y, \sigma(z_{S})) \\
    &+ \alpha T^2 \cdot KL(\sigma(z_{T}/T),\sigma(z_{S}/T)),
\end{aligned}
\end{split}
\label{eq:KD}
\end{equation}
where $L_{KL}$ denotes the KL-divergence function, $\sigma(\cdot)$ denotes the softmax function, $z_{S}$ and $z_{T}$ are the output logit vectors of the student and the teacher model, respectively. 
$T$ and $\alpha$ denotes the hyperparameter temperature and balancing weight.
Note that we can combine Hinton's knowledge distillation with our method by simply changing $L_{CE}$ to $L_{KD}$ in Equation \ref{eq:Loss}:
\begin{equation}
\begin{split}
\begin{aligned}
    Loss' = &(1-\alpha) \cdot L_{CE}(y, \sigma(z_{S})) \\
    & + \alpha T^2 \cdot KL(\sigma(z_{T}/T),\sigma(z_{S}/T)) + \beta\cdot L_{SK},
\end{aligned}
\end{split}
\label{eq:Loss'}
\end{equation}

%%%%%%%%%%%%%%%%%%%%%%%%%%%%%%%%%%%%%%%

\section{Experiments}
In this section, the details of a series of experiments for validating the proposed framework and the corresponding results are described and discussed. We have made the source codes available for reproducing our experiments at https://github.com/CECNL/EEG-KD.

\textbf{Dataset.}  
We validate our framework on BCI competition IV 2a Dataset (BCIC-IV-2a) \cite{brunner2008bci}, which is a commonly used motor-imagery EEG dataset.
The dataset comprises a total of 18 sessions from nine subjects, each subject was recorded in two sessions on different days, there are four classes (right hand, left hand, tongue, and feet) with 72 trials per class in one session and twenty-two electrodes were used to record the EEG.
For preprocessing, the signal is first downsampled from 256 Hz to 128 Hz, and then a 4-38 Hz bandpass filter is adopted.

\begin{figure}[!]
\centering
\captionsetup[subfloat]{labelsep=none,format=plain,labelformat=empty}
\subfloat[Four-channel headphone-like (4p) montage]
{\includegraphics[width=0.45\columnwidth,trim={4 4 4 4},clip]{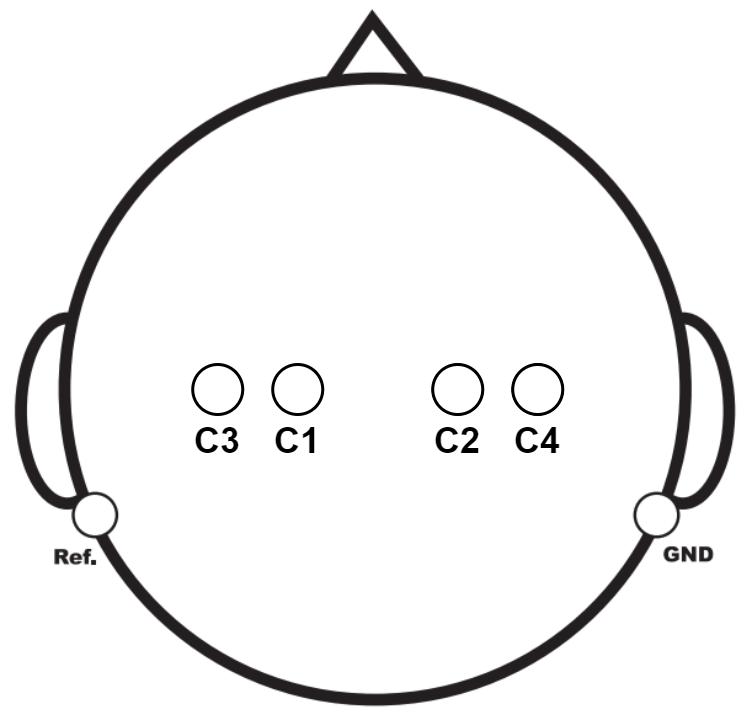}}
\hfil
\subfloat[Four-channel headband-like (4b) montage]
{\includegraphics[width=0.45\columnwidth,trim={4 4 4 4},clip]{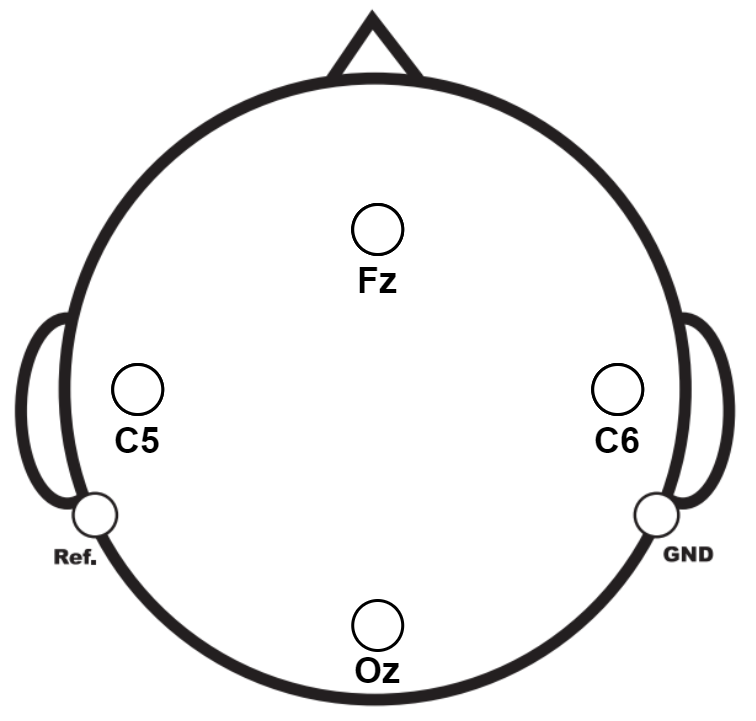}}
\\
\subfloat[Enophone, Mindset Innovation Inc. \cite{aguilar2021advanced}]
%\subfloat[Conscious Labs\protect\footnotemark]
{\includegraphics[width=0.45\columnwidth]{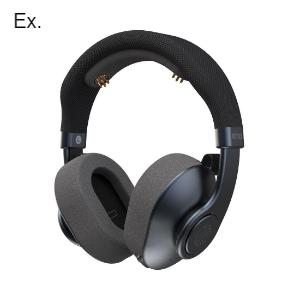}}
\hfil
\subfloat[FocusCalm, BrainCo Inc. \cite{FocusCalm}]
%\subfloat[InteraXon\protect\footnotemark]
{\includegraphics[width=0.45\columnwidth]{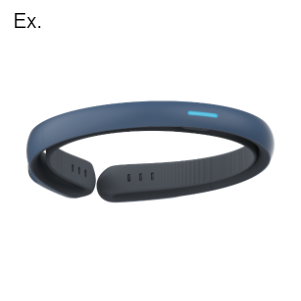}}
\caption{Low-density EEG montages and the examples of their corresponding wearable devices. Note that the number of electrodes on the example device may differ from our montages.}
\label{fig:Montage}
\end{figure}

\textbf{Montage.}
Low-density EEG devices are often designed as headphone-like or headband-like wearables. 
Available examples include Enophone (Mindset Innovation Inc.) \cite{aguilar2021advanced}, and FocusCalm EEG headband (BrainCo Inc.) \cite{FocusCalm}.
As shown in Fig. \ref{fig:Montage}, we chose the two 4-electrode montages to mimic the montages of headphone-like and headband-like low-density EEG devices.
% In our study, we choose a 7-electrode montage to mimic the montage of a headphone-like low-density EEG device, and a 4-electrode montage for a headband-like device as shown in Fig. \ref{fig:Montage}.

\textbf{Neural network architecture for EEG decoding.}
We employed representative EEG decoding neural networks, ShallowConvNet \cite{schirrmeister2017deep}, EEGNet \cite{lawhern2018eegnet}, and SCCNet \cite{wei2019spatial}, that have been commonly used previously studies. 
The input sizes of teacher/student models are adapted to the input EEG data size of high/low-density montages. 
The EEG latent feature maps for computing the SK loss can be collected from different layers in the network models and refer to different semantics of EEG feature extraction.
We assigned Feature 1-3 for the feature maps extracted from the upstream to downstream layers of the representative models, and their semantics are summarized as in Table \ref{tab:features}.

\begin{table}[!]
\centering
    \resizebox{\columnwidth}{!}{%
\begin{tabular}{c|c|c|c}
\toprule
Network  & SCCNet & EEGNet & ShallowConvNet    \\ \midrule
LF1  & Spatial & Temporal & Temporal \\ \midrule
LF2  & Spatial-temporal & Temporal-spatial & Temporal-spatial\\ \midrule
LF3  & Power        & Temporal-spatial & Power \\ \bottomrule
\end{tabular}
}
    \caption{Semantics of feature maps in the EEG decoding neural network models. LF1-LF3 refers to the latent feature maps extracted from the upstream, midstream, and downstream layers of the model.}
    \label{tab:features}
\end{table}

\textbf{Implementation details.} 
We used PyTorch to construct our experiment environment. 
The total training epoch is $500$, the optimizer is Adam optimizer \cite{kingma2014adam}, the leaning rate is $0.0005$, the weight decay is $0.1$, and the mini-batch size is $128$.
The training set and the test set were divided according to the subject-individual scheme \cite{wei2019spatial}, where the first session of a subject was the training set and the second session was the test set.
To avoid overfitting, we used $1/8$ of the training set to form the validation set that has an equal number of trials for each class and save the network at the epoch where the lowest validation loss was obtained.
We set $\alpha$ as $0.9$, T as $4$, and $\beta$ as $450$, and other hyperparameters follow the setting reported in their own research.
We used $Net-n$ to denote the neural network with $n$ electrodes as input, and $p$ and $b$ to denote the 4-channel headphone-like (4p) and headband-like (4b) montages, respectively.
All of the student model performances were averaged over 10 repeated runs with random initialization.

\begin{table*}
  \centering
\resizebox{\textwidth}{!}{%
\begin{tabular}{
 c| *{5}{S[table-format=-2.2,table-space-text-post=***]|}
    *{1}{S[table-format=-2.2,table-space-text-post=***]}}
\toprule

\multirow{2}{*}{Teacher} & {SCCNet-22} & {SCCNet-22} & {EEGNet-22} & {EEGNet-22} & {ShallowConvNet-22} & {ShallowConvNet-22} \\
& 80.25 & 80.25 & 67.82 & 67.82 & 74.61 & 74.61 \\
\midrule
\multirow{2}{*}{Student} & {SCCNet-4p} & {SCCNet-4b} & {EEGNet-4p} & {EEGNet-4b} & {ShallowConvNet-4p} & {ShallowConvNet-4b} \\
& 52.13 & 47.38 & 46.37 & 44.15 & 50.32 & 43.81 \\
\midrule
LF1     & 53.38 & 48.99 & 46.99 & 46.27 & 47.18 & 45.23 \\
LF2     & 54.80 & 50.47 & 48.65 & 44.21 & 52.13 & 46.45 \\
LF3     & 55.57 & 49.83 & 51.89 & 47.23 & 51.78 & 46.69 \\
\midrule
LF1+2   & 54.75 & 50.82 & 47.83 & 45.98 & 49.16 & 46.63 \\
LF2+3   & 55.31 & 50.84 & 51.93 & 46.47 & 52.33 & 47.43 \\
LF1+2+3 & 54.83 & 51.32 & 49.11 & 46.89 & 49.11 & 46.93 \\
\bottomrule
\end{tabular}}

  \caption{Accuracy (\%) of the proposed similarity-keeping distillation using different types of feature maps used for estimation of similarity matrices.}
  \label{tab:layers}
\end{table*}

\subsection{Feature maps for distillation}
% \textbf{Feature maps for distillation.}
As mentioned above, the feature maps at different layers within a neural network model associate with different semantics. 
We first conducted an experiment to study the influence of using a single feature map or using multiple feature maps at different depths simultaneously in our framework, and the results are shown in Table \ref{tab:layers}. 
We observe that when only one feature map is distilled, student networks with high-level (downstream) feature maps outperform others in most cases, and this result supports the statement above that deeper feature maps are more suitable for transferring knowledge.
When combining different feature maps, adding high-level feature maps mostly provides further improvement.
On the contrary, the addition of low-level feature maps does not always yield better results, and we assume that it is because regularization for low-level features may restrict the flexibility of student networks \cite{romero2014fitnets}.
Feature maps at downstream layers in a neural network integrate the information from previous layers to generate more discriminative high-level features that can be used to classify input samples, and most relation-based KD methods use only the representations at the penultimate layer (prior to logits) to construct structural knowledge to be transferred \cite{tian2019contrastive,tung2019similarity,passalis2018probabilistic}.
Although SCCNet-4b and EEGNet-4b acquire higher improvements using all feature maps, we used feature 2+3 for a stable overall performance across architectures in our framework.

\begin{table*}[!ht]
  \begin{threeparttable}
  \centering
\resizebox{\textwidth}{!}{%
\begin{tabular}{
  c| *{3}{S[table-format=-2.2,table-space-text-post=***]}|
     *{3}{S[table-format=-2.2,table-space-text-post=***]}|
     *{3}{S[table-format=-2.2,table-space-text-post=***]}
}
\toprule
\multirow{2}{*}{Teacher} 
& & {SCCNet-22} &
& & {EEGNet-22} &
& & {ShallowConvNet-22} & \\
& & 80.25 & & & 67.82 & & & 74.61 & \\
\midrule
\multirow{2}{*}{Student} 
& {SCCNet-4p} & {EEGNet-4p} & {ShallowConvNet-4p}
& {SCCNet-4p} & {EEGNet-4p} & {ShallowConvNet-4p} 
& {SCCNet-4p} & {EEGNet-4p} & {ShallowConvNet-4p}\\
& 52.13 & 46.37 & 50.32 
& 52.13 & 46.37 & 50.32 
& 52.13 & 46.37 & 50.32 \\
\midrule
  HKD &
54.39 & \color{red} 48.25* & 51.08 & 53.66 & \color{red} 48.83** & 49.59 & 54.11 & 47.42 & 51.15 \\
FitNet &
    50.83 & \color{red} 49.66** & \color{blue} 45.47** & 50.90 & 47.35 & \color{blue} 43.76** & \color{blue} 46.20* & 43.60 & \color{blue} 46.50** \\
AT &
  54.31 & \color{red} 51.39** & \color{red} 53.12* & 54.53 & \color{red} 50.22* & 52.09 & 53.61 & \color{red} 51.40** & 51.38 \\
VID &
  53.33 & 46.93 & 50.12 & 53.79 & \color{red} 47.30* & 50.75 & 53.63 & 46.96 & 50.14 \\
AB &
  \color{red} 54.83* & 46.05 & 49.04 & 54.00 & \color{red} 48.95** & \color{red} 51.25* & 52.83 & 44.98 & \color{red} 51.62* \\
PKT &
  54.27 & \color{red} 48.83** & \color{red} 51.65** & 53.83 & \color{red} 51.18** & \color{red} 51.83* & \color{red} 54.02* & \color{red} 47.78* & 50.79 \\
CC &
  53.34 & 46.41 & 50.55 & 53.55 & 46.80 & 50.64 & 52.93 & 46.71 & 50.11 \\
RKD &
  54.32 & \color{red} 48.39** & 51.90 & 53.90 & \color{red} 48.98* & \color{red}  52.23* & 54.11 & 47.85 & 50.79 \\
SP &
  \color{red} 54.11* & 48.08 & 52.35 & 52.22 & \color{red} 51.84** & 49.75 & 54.01 & 46.94 & 51.54 \\
CRD &
  53.98 & 48.47 & \color{blue} 49.24** & 53.25 & \color{red} 48.34* & 48.65 & 53.29 & 48.01 & 49.67 \\ \midrule
FitNet (w/ logits loss) &
  51.98 & \color{red} 50.07** & 48.48 & 51.15 & 47.27 & 46.82 & 48.70 & 45.24 & \color{blue} 47.74* \\
AT (w/ logits loss) &
  54.61 & \color{red} 51.18** & \color{red} 53.01* & \B 54.75 & 50.03 & \B 52.43 & 54.48 & \color{red} 50.37** & 51.55 \\
VID (w/ logits loss) &
  54.31 & \color{red} 47.89* & 51.33 & 53.60 & \color{red} 49.10** & 48.89 & 54.21 & 47.47 & 51.07 \\
AB (w/ logits loss) &
  54.54 & 47.20 & 50.57 & 53.61 & \color{red} 50.66** & 51.56 & 53.67 & 46.22 & \color{red} 52.45* \\
PKT (w/ logits loss) &
  54.47 & \color{red} 49.43** & 51.52 & 53.50 & \color{red} 50.94* & 50.35 & 53.98 & 48.06 & 51.15 \\
CC (w/ logits loss) &
  54.28 & \color{red} 47.96* & 50.34 & 53.16 & \color{red} 49.32** & 48.64 & 53.80 & 47.70 & 50.25 \\
RKD (w/ logits loss) &
  \color{red} 54.94* & \color{red} 49.02** & 52.03 & 53.65 & \color{red} 50.73** & 50.86 & 54.45 & \color{red} 48.38* & 51.60 \\
SP (w/ logits loss) &
  \color{red} 54.61* & \color{red} 49.99** & 52.36 & 52.36 & \color{red} 51.66** & 50.15 & 54.01 & \color{red} 49.16* & 51.83 \\
CRD (w/ logits loss) &
  54.49 & \color{red} 50.05** & 51.12 & 53.14 & 49.06 & 49.19 & 54.18 & \color{red} 49.24** & 50.95 \\ \midrule  
SK &
  \color{red} \B  55.31* & \color{red} 52.52** & \color{red} \B  53.17* & 54.41 & \color{red} \B 51.93** & 52.23 & \color{red} \B 54.88** & \color{red} \B 51.60** & \color{red} 52.33** \\
 SK (w/ logits loss) &
  \color{red} 55.20* & \color{red} \B  53.41** & 53.01 & 53.36 & \color{red} 51.76** & 51.46 & 54.88 & \color{red} 51.05** & \color{red} \B 52.75*
  \\ \bottomrule
\end{tabular}
}

  \begin{tablenotes}[flushleft]
    \item 
    \tiny{\textbf{Bold}: Highest accuracy.}
    \tiny{{\color{red} Red}: Significantly outperform baseline student networks.}
    \tiny{{\color{blue} Blue}: Significantly underperform baseline student networks.}
    \item
    \tiny{*$p<$0.05, **$p<$0.01, ***$p<$0.001}
  \end{tablenotes}
  \end{threeparttable}
  \caption{Accuracy (\%) of student networks across knowledge distillation methods on the headphone-like montage.}
  \label{tab:7woKD}
\end{table*}

\begin{table*}
    \begin{threeparttable}
      \centering
\resizebox{\textwidth}{!}{%
\begin{tabular}{
  c| *{3}{S[table-format=-2.2,table-space-text-post=***]}|
     *{3}{S[table-format=-2.2,table-space-text-post=***]}|
     *{3}{S[table-format=-2.2,table-space-text-post=***]}
}
\toprule
\multirow{2}{*}{Teacher} 
& & {SCCNet-22} &
& & {EEGNet-22} &
& & {ShallowConvNet-22} & \\
& & 80.25 & & & 67.82 & & & 74.61 & \\
\midrule
\multirow{2}{*}{Student} 
& {SCCNet-4b} & {EEGNet-4b} & {ShallowConvNet-4b}
& {SCCNet-4b} & {EEGNet-4b} & {ShallowConvNet-4b} 
& {SCCNet-4b} & {EEGNet-4b} & {ShallowConvNet-4b}\\
& 47.38 & 44.15 & 43.81 
& 47.38 & 44.15 & 43.81 
& 47.38 & 44.15 & 43.81 \\
\midrule
  HKD &
  48.91 &
  \color{red} 45.69* &
  \color{red} 46.78* &
  48.93 &
  45.15 &
  \color{red} 46.76* &
  \color{red} 49.37* &
  45.37 &
  \color{red} 48.05** \\
FitNet &
  48.95 &
  \color{red} 48.37** &
  45.49 &
  47.35 &
  45.35 &
  44.93 &
  \color{blue} 45.35* &
  42.97 &
  \color{red} 47.55* \\
AT &
  47.42 &
  44.10 &
  43.64 &
  \color{red} 47.58* &
  44.07 &
  44.17 &
  47.38 &
  44.02 &
  44.10 \\
VID &
  \color{red} 48.09* &
  44.05 &
  44.75 &
  \color{red} 48.02* &
  43.82 &
  45.11 &
  \color{red} 48.03* &
  43.88 &
  45.52 \\
AB &
  \color{red} 48.66* &
  43.67 &
  42.91 &
  47.64 &
  44.28 &
  45.74 &
  46.26 &
  \color{blue} 41.82* &
  \color{red} 46.55* \\
PKT &
  \color{red} 49.09** &
  45.22 &
  \color{red} 47.09** &
  49.02 &
  \color{red} 47.00* &
  \color{red} 47.47* &
  48.35 &
  45.11 &
  \color{red} 45.54* \\
CC &
  \color{red} 49.51* &
  44.32 &
  \color{red} 47.97* &
  \color{red} 49.11* &
  44.19 &
  \color{red} \B 47.89* &
  \color{red} 49.23* &
  44.54 &
  \color{red} 47.90* \\
RKD &
  46.05 &
  44.03 &
  44.68 &
  45.85 &
  44.75 &
  44.53 &
  46.03 &
  45.10 &
  44.46 \\
SP &
  \color{red} 50.45* &
  43.82 &
  \color{red} 49.08* &
  48.01 &
  \color{red} 47.10* &
  46.41 &
  49.49 &
  43.31 &
  47.93 \\
CRD &
  45.54 &
  \color{blue} 41.39** &
  43.81 &
  45.64 &
  43.57 &
  43.85 &
  \color{blue} 44.99* &
  \color{blue} 41.90** &
  42.79 \\ \midrule
FitNet (w/ logits loss) &
  \color{red} 49.30* &
  \color{red} \B  49.67** &
  \color{red} 46.84* &
  48.65 &
  45.97 &
  45.08 &
  47.74 &
  45.97 &
  47.06 \\
AT (w/ logits loss) &
  48.91 &
  \color{red} 45.74* &
  \color{red} 46.79* &
  49.02 &
  45.03 &
  \color{red} 46.72* &
  \color{red} 49.38* &
  45.36 &
  \color{red} 48.03** \\
VID (w/ logits loss) &
  \color{red} 49.35* &
  \color{red} 45.86* &
  \color{red} 47.42** &
  48.88 &
  45.17 &
  \color{red} 47.26* &
  \color{red} 49.79* &
  45.17 &
  \color{red} 48.81** \\
AB (w/ logits loss) &
  \color{red} 49.83** &
  \color{red} 46.66** &
  45.51 &
  47.64 &
  44.28 &
  \color{red} 47.32* &
  \color{red} 49.37** &
  43.78 &
  \color{red} \B  49.00** \\
PKT (w/ logits loss) &
  \color{red} 49.85** &
  \color{red} 47.09** &
  \color{red} 47.40* &
  49.17 &
  46.81 &
  \color{red} 46.92* &
  \color{red} 49.78* &
  \color{red} 46.55** &
  \color{red} 48.54** \\
CC (w/ logits loss) &
  \color{red} 50.04** &
  \color{red} 45.86* &
  \color{red} 48.71** &
  48.98 &
  45.27 &
  \color{red} 47.59* &
  \color{red} 49.87* &
  \color{red} 45.72* &
  \color{red} 48.60** \\
RKD (w/ logits loss) &
  \color{red} 49.97* &
  \color{red} 46.27* &
  \color{red} 48.74* &
  48.47 &
  45.10 &
  \color{red} 47.92* &
  49.52 &
  44.39 &
  \color{red} 48.10* \\
SP (w/ logits loss) &
  \color{red} 50.74** &
  46.12 &
  \color{red} \B  49.42* &
  49.15 &
  \color{red} \B  47.15* &
  \color{red} 46.72* &
  \color{red} 50.00* &
  45.76 &
  \color{red} 48.97** \\
CRD (w/ logits loss) &
  48.36 &
  44.88 &
  \color{red} 46.84* &
  47.43 &
  44.98 &
  \color{red} 46.59* &
  48.96 &
  45.03 &
  \color{red} 47.33* \\ \midrule  
SK &
  \color{red} 50.84** &
  \color{red} 48.44** &
  \color{red} 48.56* &
  \color{red} \B 49.61* &
  46.47 &
  \color{red} 47.69* &
  \color{red} 49.23* &
  \color{red} \B 48.40** &
  \color{red} 47.43* \\
 SK (w/ logits loss) &
  \color{red} \B  51.26** &
  \color{red} 49.35** &
  \color{red} 49.34** &
  49.07 &
  45.86 &
  \color{red} 47.26* &
  \color{red} \B  50.34* &
  \color{red} 48.17** &
  \color{red} 48.36*
  \\ \bottomrule
\end{tabular}
}

      \begin{tablenotes}[flushleft]
        \item 
        \tiny{\textbf{Bold}: Highest accuracy.}
        \tiny{{\color{red} Red}: Significantly outperform baseline student networks.}
        \tiny{{\color{blue} Blue}: Significantly underperform baseline student networks.}
        \item
        \tiny{*$p<$0.05, **$p<$0.01, ***$p<$0.001}
      \end{tablenotes}
      \end{threeparttable}
  \caption{Accuracy (\%) of student networks across knowledge distillation methods on the headband-like montage.}
  \label{tab:4woKD}
\end{table*}

\subsection{Performance comparison}

This section presents the performance of the student models in motor-imagery EEG classification with KD methods across model architectures when headphone-like and headband-like montages are used.

Table~\ref{tab:7woKD} summarizes the performance of the headphone-like student models and compares our proposed SK framework against other leading KD methods (Hinton's KD is denoted as HKD for simplicity). 
The student model performance without the use of KD is employed as the baseline for evaluating the improvement gained from individual KD methods. 
The improvement was tested by the Wilcoxon signed rank test for the significance level. 
Bold accuracies denote the highest performance of student network models in each column.
Red and blue accuracies denote the significant positive improvement and significant negative deterioration, and the asterisk denotes the significance level based on the p-value.
The best headphone-like student model performance is with our SK framework regardless of the use of logits loss. 
Meanwhile, the greatest improvement of 7.04\% can be provided by our method with the pair of SCCNet-22 (teacher) \& EEGNet-4p (student).
We observe that our SK method consistently improves the accuracies of student networks and significantly outperforms the baseline student model in most cases.
Although our method does not always have the highest accuracy in all combinations of teacher and student network models, our method can still obtain comparable results in those cases.
In addition, most competing methods do not support stable improvement, and PKT is the only method that can stably improve student models among them, while our method can provide higher improvement than PKT.
Our results suggest that competing methods may not be suitable for our task because of the information gap caused by the density difference between their montages, which is not considered when they are proposed.
No observable differences are reported on the effectiveness of feature-based and relation-based KD methods.
Nevertheless, the only two KD methods that offer consistent increase of accuracy across all architectures are both relation-based methods (PKT and our SK).

The performance of the headband-like student model is shown in Table~\ref{tab:4woKD}. 
The baseline student model accuracies are obviously lower than those with the headphone-like montage, suggesting the information carried in the 4-electrode low-density EEG data may be less sufficient in the task of motor-imagery EEG classification.
An interesting finding is that, on the headband-like montage, our method shows less superiority, and there are fewer cases for competing methods to significantly underperform baseline student models.
One possible explanation might be that it is difficult for student network models to completely learn from the teacher model, so the difference between methods would be eliminated.
Furthermore, limited information is an obstacle to learning discriminative features, so these competing methods would also be helpful as additional regularizations.

As our SK framework incorporates both the SK loss and the logits loss (Hinton's KD), we also validate the compatibility of logits loss with our SK loss and other KD methods for a fair comparison.
The results are shown at the bottom of Tables \ref{tab:7woKD} and \ref{tab:4woKD}.
%(Logits / ours)
Unfortunately, combining Hinton's KD with our method does not provide a consistent additional improvement.
%(Logits / other methods)
In contrast, combining with Hinton's KD benefits other KD methods compensates for most decrease in performance.
A surprising finding is that all KD methods outperform baseline student networks with the help of logits loss on the headband-like montage.
These results suggest that Hinton's KD loss should be added to the total loss function when distilling an EEG network, and the source and target tasks are the same.

Another observation is that the effect of our method depends on the quality of the teacher network.
Specifically, with a better teacher network, our method can further enhance the improvement of the student network.
For instance, when we used SCCNet-22 as the teacher network, the student network could surpass those using EEGNet-22 or ShallowConvNet-22 as the teacher network.
However, we also see performance improvement tends to depend on the choice of student network model in other suboptimal cases, such as those improvements with using EEGNet-4p.

\begin{figure*}[!ht]
\centering
	\subfloat{\includegraphics[width=0.49\textwidth,trim=2 2 2 2,clip]{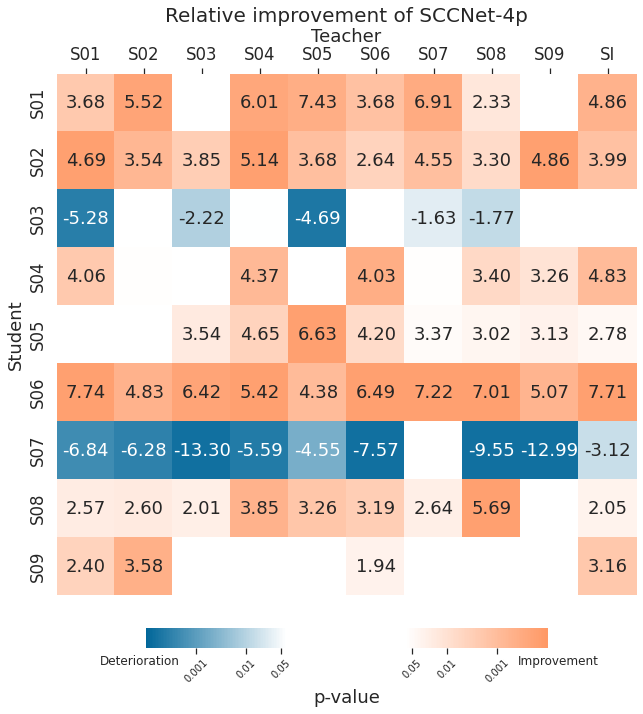}}
	\hfill
	\subfloat{\includegraphics[width=0.49\textwidth,trim=2 2 2 2,clip]{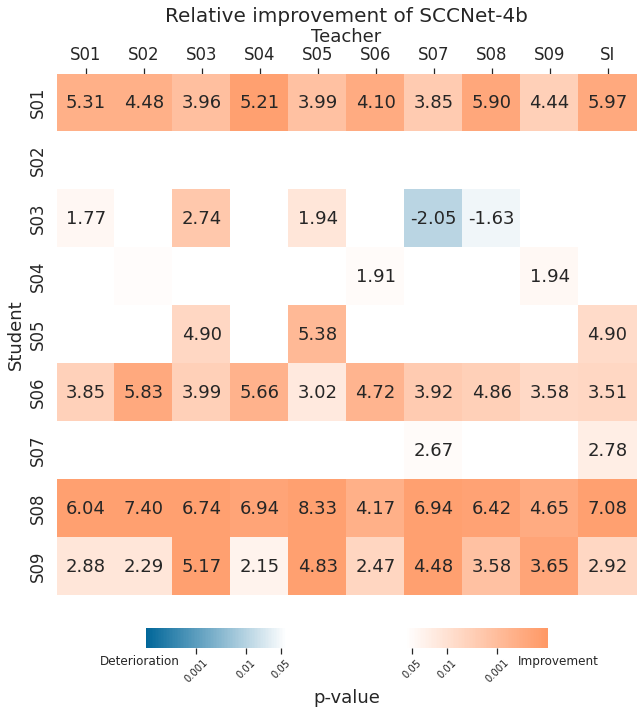}}
\caption{Performance improvement by using the proposed SK framework when the teacher and student models are trained across subjects. 
    The experiments used the model architecture, SCCNet-22, that offers the highest decoding performance according to previous evaluations. 
    Two heatmaps are corresponding to the SCCNet-7 and SCCNet-4 for the (left) 7-electrode montage and (right) 4-electrode montage used in our study. 
    The teacher model is trained by different subjects (S01-S09) and all other subjects (SI) with respect to the target subject using a student model.
    The Wilcoxon rank-sum test is used to test the significance level of the differences to baseline student networks, and we only show relative performance for those with statistically significant difference.}
\label{fig:Cross-subject}
\end{figure*}

\subsection{Distillation across subjects}
Although our SK framework requires a teacher model to improve the performance of a low-density student EEG decoding model, it is not necessary to train the teacher model with the same subject's data with the student model. 
In this part, we present the experimental results in the cross-subject teacher-student KD scenario.

We investigated the effect of distilling knowledge from the pre-trained high-density network models of different subjects to the low-density networks for a target subject, and Fig. \ref{fig:Cross-subject} depicts the relative performance improvement of different montages for student network models compared to the baseline.
Overall, we observe variaiblity of relative improvement given by our SK framework in both headphone-like and headband-like montages. 

Student networks for most subjects can be improved with the help of teacher networks of different subjects.
Although there are still individual differences between subjects, these results show that our KD framework can be used across different subjects, and we may skip the teacher network pre-training process when high-density EEG networks of other subjects are available.

In addition, we followed the subject-independent scheme in \cite{wei2019spatial} to train new teacher network models whose training data are from all other subjects (for example, the subject-independent network for S01 is trained with the data from S02 to S09), and most of these SI teacher network models can also improve student models.
However, those SI teacher models that integrated data from multiple subjects, did not always outperform other teacher network models.
This result suggests that further research is needed to seek an optimal scheme to leverage data from multiple subjects when distilling knowledge of EEG data.

\subsection{Number of electrodes}

\begin{figure}[!]
    \centering
    \includegraphics[width=\columnwidth]{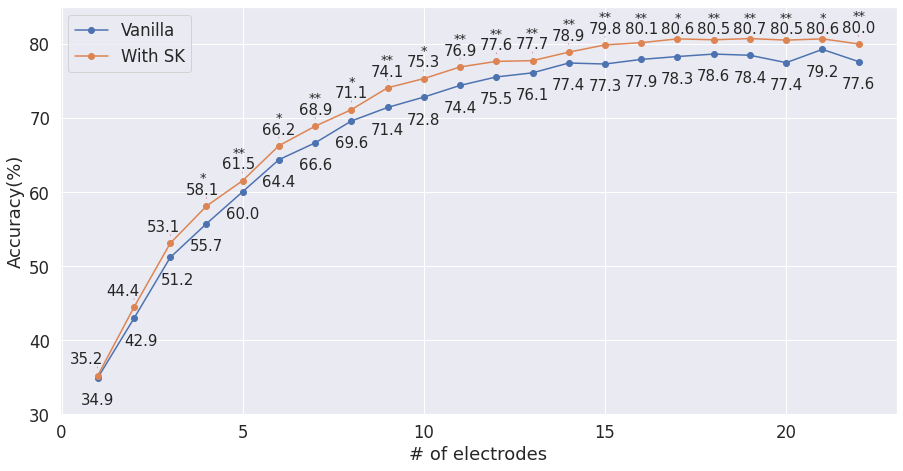}
    \caption{
        Comparison of vanilla student model (w/o KD) and models trained with our SK loss under different number of electrodes.
        The Wilcoxon rank-sum test is used to test the significance level of the differences, and the significance levels are shown above the performance of SK (*$p<$0.05, **$p<$0.01, ***$p<$0.001).
    }
    \label{fig:number}
\end{figure}

\begin{figure}[!]
    \centering
    \includegraphics[width=\columnwidth]{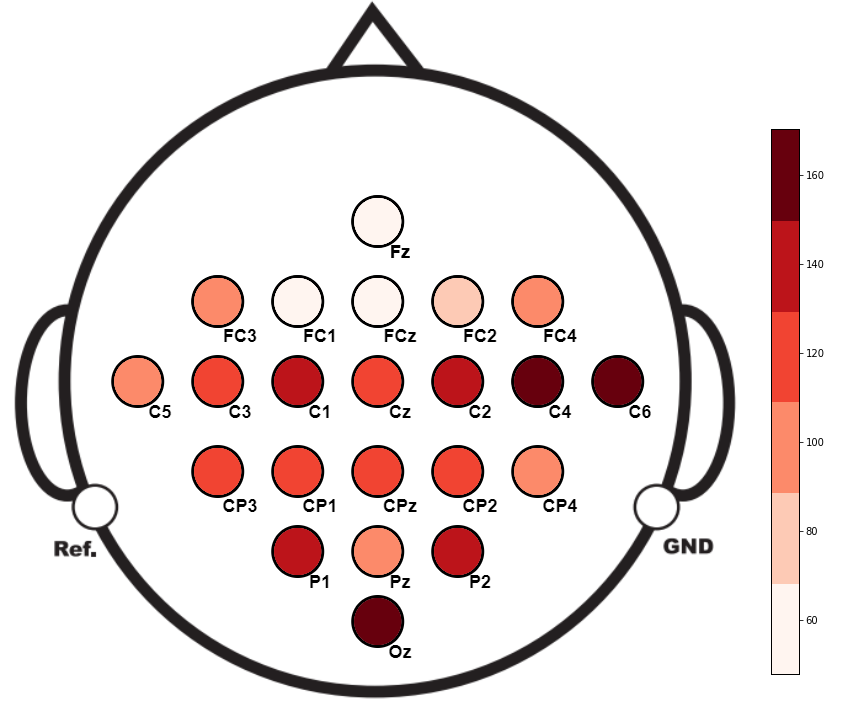}
    \caption{
        Frequency of usaged in the backward feature selection of each electrode. 
        Dark red marks a higher frequency.
    }
    \label{fig:important}
\end{figure}

Here, we discuss the influence of the number of input electrodes on the effectiveness of our SK framework, and the results are shown in Fig. \ref{fig:number}.
We applied backward feature selection where electrodes were sorted according to the decrease in performance as each of them was excluded and then sequentially removed the input electrodes of models.
Our feature selection is similar to recursive feature elimination \cite{guyon2002gene}, where the least important feature is removed at the beginning of the recursion.
Interestingly, our SK framework can provide consistent improvement regardless of the number of input electrodes.
Even when sufficient electrodes are used, our SK framework can still enhance the generalizability of student models.
These results suggest that our framework is not limited to improving decoding performance for low-density EEG models but has the potential for other problems where transferring information across channels is needed.

Fig. \ref{fig:important} illustrates the importance score for each electrode.
Scores are calculated by summing up the frequencies of use during the above training process.
C4, C6, and Oz channels have the highest usage, while Fz, FC1, and FC2 channels are eliminated early more often. The
The low importance score of Fz might be the reason why the average performance of the headband-like montage is worse than that of the headphone-like montage.

\subsection{Ablation study}

\begin{table}[t!]
    \resizebox{\columnwidth}{!}{%
\begin{tabular}{cccccc}
\multicolumn{6}{c}{ \small{w/ zero-centered feature maps}}  \\ \toprule
     & Baseline & L2-norm         & PLV       & Dot product        & Ours     \\ \midrule
Acc  & 52.13    & 44.48      & 53.51     & 54.82     & 55.31    \\
Diff & -        & -7.65      & +1.37     & +2.69     & +3.18    \\ \bottomrule
\\
\multicolumn{6}{c}{ \small{w/o zero-centered feature maps}} \\ \toprule
     & Baseline & L2-norm         & PLV       & Dot product        & Ours     \\ \midrule
Acc  & 52.13    & 43.58      & 53.61     & 54.11     & 55.12    \\
Diff & -        & -8.55      & +1.47     & +1.98     & +2.99    \\ \bottomrule
\end{tabular}%
}
    \caption{Ablation study for the key components of the SK loss estimation.}
    \label{tab:ablation}
\end{table}

An ablation experiment was carried out to assess the effectiveness of the components in our SK loss function, which calculates the cosine similarity between the zero-centered feature maps.
Table~\ref{tab:ablation} shows the results of using different similarity criteria with or without zero-centered feature maps, and SCCNet-22 and SCCNet-4p are employed as teacher and student networks.
$L_{2}$ directly calculates the Euclidean distance between the feature maps, however, both versions with/without zero-centered feature maps fail to provide an improvement.
PLV denotes the phase locking value \cite{aydore2013note}, which can assess the synchronization between the phases of two input signals.
The dot product was used in the SP method \cite{tung2019similarity} as a metric of latent feature similarity.
Our method provides the highest improvement of 3.18\% with zero-centering and cosine similarity, while the version without zero-centering feature maps shows a 2.99\% improvement.
These results evident the effectiveness of the cosine similarity-based similarity-keeping mechanism as well as the benefit of zero-centered feature maps.
Further exploration may target to incorporating domain adaptation techniques to deal with the domain shift problem raised from transferring knowledge of models across multiple subjects/sessions \cite{hang2019cross,li2022subject}.

%%%%%%%%%%%%%%%%%%%%%%%%%%%%%%%%%%%%%%%

\section{Conclusion}
In this work, we propose the first KD framework dedicated to transferring knowledge across EEG decoding models. The experimental results suggest consistent improvement on low-density EEG decoding performance using the proposed SK method that captures inter-sample similarity within the latent domain. Furthermore, the validation of our SK method on cross-subject KD suggests the practicality in EEG-based BCI with low-density wearable devices in a real-world context.
We foresee our proposed SK framework to be applied in various scenarios where EEG decoding performance can be improved by auxiliary knowledge transferred from pre-trained high-performance models.

\bibliographystyle{IEEEtran}
\bibliography{EEG_KD_v0}
% \bibliography{Mendeley}

\end{document}